%% file: bare_jrnl_new_sample4.tex
\documentclass[lettersize,journal]{IEEEtran}
\usepackage{amsmath,amsfonts}
\usepackage{algorithmic}
\usepackage{algorithm}
\usepackage{array}
\usepackage[caption=false,font=normalsize,labelfont=sf,textfont=sf]{subfig}
\usepackage{textcomp}
\usepackage{stfloats}
\usepackage{url}
\usepackage{verbatim}
\usepackage{booktabs}
\usepackage{multirow}
\usepackage{makecell}
\usepackage{graphicx}
\usepackage{cite}

\begin{document}

\title{MagicAnime: A Hierarchically Annotated, Multimodal and Multitasking Dataset with Benchmarks for Cartoon Animation Generation}

\author{
    Shuolin Xu$^{\dagger 1}$, 
    Bingyuan Wang$^{\dagger 2}$, 
    Zeyu Cai$^{2}$, 
    Fangteng Fu$^{2}$, 
    Yue Ma$^{3}$ \\ 
    Tongyi Lee$^{5}$~\IEEEmembership{Senior Member,~IEEE}, 
    Hongchuan Yu$^{1*}$, 
    Zeyu Wang$^{2,3*}$%
    \thanks{$^{\dagger}$ Equal contribution.}%
    \thanks{* Corresponding author.}%
    \\
    $^1$ National Centre for Computer Animation, Bournemouth University \\
    $^2$ Hong Kong University of Science and Technology (Guangzhou) \\
    $^3$ Hong Kong University of Science and Technology \\
    $^5$ Department of Computer Science and Information Engineering, National Cheng Kung University
}




\maketitle

\input{sec/0_abstract}


\input{sec/1_introduction}

\input{sec/2_related_work}
\input{sec/3_method}
\input{sec/4_evaluation}

\input{sec/5_conclusion}



 




\vfill

\end{document}

%% file: sec/0_abstract.tex
\begin{abstract}

Generating high-quality cartoon animations
multimodal control is challenging due to the complexity of
non-human characters, stylistically diverse motions and fine-grained emotions. There is a huge domain gap between real-world videos and cartoon animation, as cartoon animation is usually abstract and has exaggerated motion. Meanwhile, public multimodal cartoon data are extremely scarce due to the difficulty of large-scale automatic annotation processes compared with real-life scenarios. To bridge this gap, We propose the MagicAnime dataset, a large-scale, hierarchically annotated, and multimodal dataset designed to support multiple video generation tasks, along with the benchmarks it includes. Containing 400k video clips for image-to-video generation, 50k pairs of video clips and keypoints for whole-body annotation, 12k pairs of video clips for video-to-video face animation, and 2.9k pairs of video and audio clips for audio-driven face animation. Meanwhile, we also build a set of multi-modal cartoon animation benchmarks, called MagicAnime-Bench, to support the comparisons of different methods in the tasks above. Comprehensive experiments on four tasks, including video-driven face animation, audio-driven face animation, image-to-video animation, and pose-driven character animation, validate its effectiveness in supporting high-fidelity, fine-grained, and controllable generation. 
\end{abstract}


%% file: sec/1_introduction.tex
\begin{figure*}[ht!]
    \centering
    \includegraphics[width=1.0\textwidth]{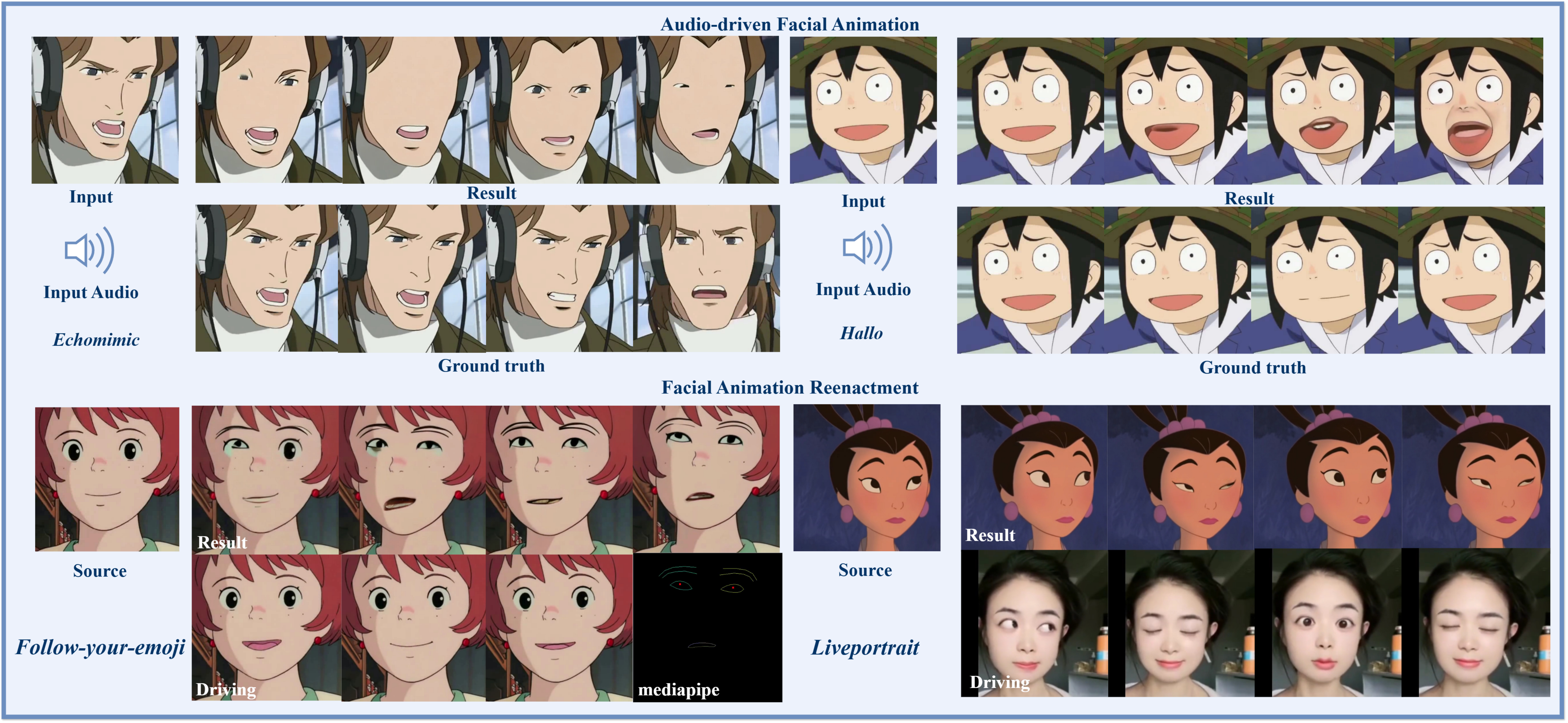}
    \caption{Various failure cases on cartoon animation generation tasks: (a) Image2Video on these methods~\cite{xing2024tooncrafter,xing2023dynamicrafter,Vondrick2016GeneratingVW} we evaluated the qualitative performance used our Image2Video test set including 300 video clips. (b) Audio2Portrait test on these methods~\cite{wei2024aniportrait,chen2024echomimic,xu2024hallo}, used our Audio2Portrait test containing 100 portrait video clips and aligned audio. (c) We used our portrait test set to do a self-reenactment test on these methods~\cite{MooreThreads_Moore_AnimateAnyone,guo2024liveportrait,ma2023follow}.}
    \label{fig:original}
    \vspace{-0.35cm}
\end{figure*}

\begin{figure*}[t!]
    \centering
    \includegraphics[width=\textwidth]{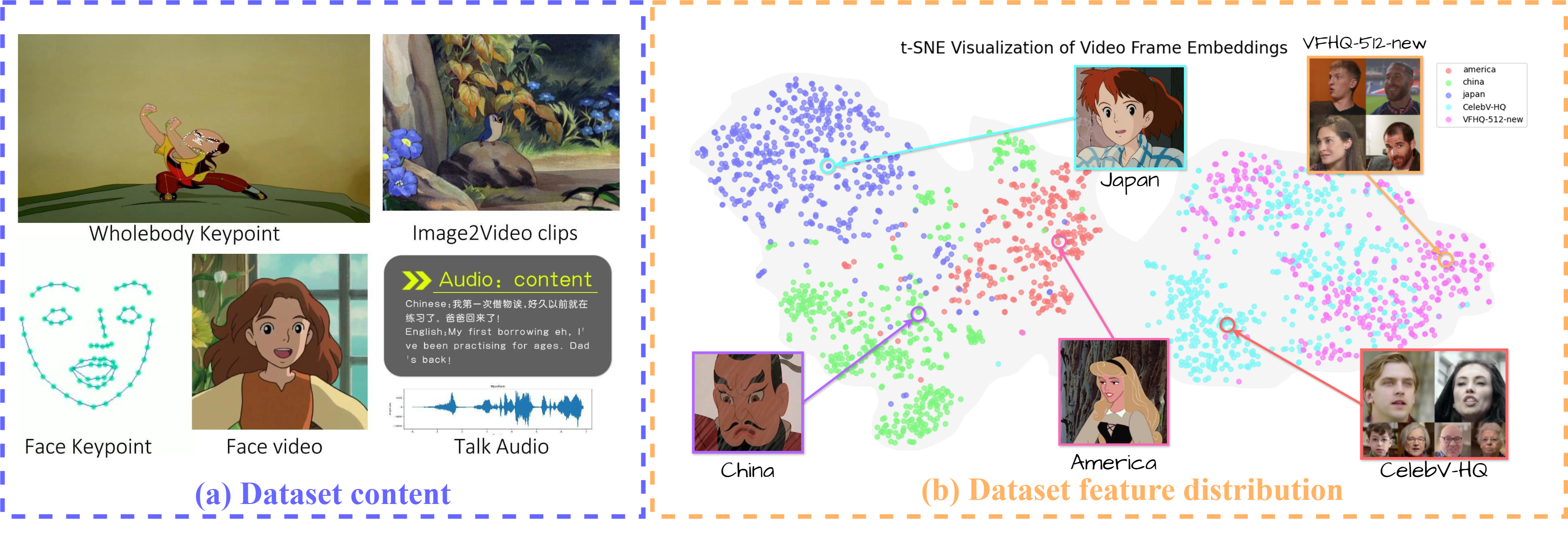}
    \caption{Dataset overview: (a) Dataset modalities. Our dataset consists of different modalities including whole body and face keypoint frame-by-frame, video, audio, \textit{etc}. (b) Dataset feature distribution. We extract the first frame of each video clip and conduct t-SNE dimension reduction of their CLIP embeddings. The results demonstrate independent and diverse feature distribution of our constructed dataset compared to other realistic human-centered datasets~\cite{xie2022vfhq,zhu2022celebvhq}. }
    \label{fig:tsne}
\end{figure*}

\section{Introduction}

The traditional cartoon animation industry is known for its complex workflows and substantial time and resource requirements. Recently, the rapid development of video generation technologies,
such as image-to-video (I2V) generation and text-to-video (T2V) generation~\cite{Keling,xing2023dynamicrafter,Sora,kong2024hunyuanvideo,wan2.1}, audio-driven facial animation~\cite{zhang2023sadtalker,wei2024aniportrait,xu2024hallo,cui2024hallo2}, and pose-driven animation~\cite{MooreThreads_Moore_AnimateAnyone} can be integrated into conventional cartoon animation, including lip sync, facial expression animation, character animation, and in-between animation~\cite{xing2024tooncrafter, meng2024anidoc, liu2025manganinja,zhang2024mikudance}. These automated approaches have the potential to accomplish the various tasks, introducing new opportunities for cartoon animation production.



In the context of image and video generation, most existing approaches mainly focus on the general subject domain and pre-train their model on public real-world dataset, resulting in noticeable artifacts in cartoon domain generation tasks~\cite{zhao2022cartoon}. As shown in Fig.~\ref{fig:original}, we performed experiments to evaluate the ability of SOTA in cartoon domain. Based on our observation, current methods on comic animation suffer from a series of problems including the inability to generate large-scale body motions, partial generation and color mismatches, and failure to capture accurate facial expressions and actions for the five senses. They also struggle with losing facial features, incorrect reconstruction of eyebrows and eyes, and blurring or jitter during head movements. Despite positive results in some cross-reenactment tasks~\cite{guo2024liveportrait}, there is a significant lack of effective priors from cartoon animation data.

A large number of datasets in video generation provide reference for cartoon animation generation, but most of them only a limited proportion of cartoon data. For example, Openhumanvid~\cite{li2024openhumanvid}, provides a large-scale human-centric video dataset with multiple modalities and multitask capabilities, but it primarily focuses on real-world video generation, with cartoon animation data comprising merely 1\% of the total dataset. The similar issue also exists for benchmarks. For example, VBench~\cite{huang2024vbench} is most widely used in video generation but also predominantly features real-world video content, lacking cartoon-style data. Some of its evaluation metrics, such as background consistency and motion smoothness, are also more applicable to real-world videos. Therefore, collecting a high-quality cartoon animation dataset with high-quality benchmarks is crucial and plays a vital role in the entire anime production industry.

Recent datasets such as Sakuga-42m~\cite{pan2024sakuga} and Anisora~\cite{jiang2024exploring} have alleviated the scarcity of training data and benchmarks for cartoon animation. However, they only provide textual annotation and fail to support a broader range of video generation tasks. Meanwhile, although Anisora introduced specific benchmarks for cartoon animation, there is still a lack of assessments that span multiple tasks and modalities, which is our main concern. As a result, there is a shortage of open-source datasets and benchmarks specifically tailored for the cartoon animation domain that incorporate hierarchical annotations and multimodal information. 

To fill these gaps, we present MagicAnime, a high-quality dataset and benchmark specifically designed for cartoon animation research. In our data processing pipeline, we employ hierarchical annotations, enabling both training and evaluate for a variety of tasks. Initially, we construct the core dataset based on the foundational tasks of T2V and I2V. We then expand step by step to produce subsets that support frame interpolation and pose-driven animation, ultimately deriving a facial animation subset and a more narrowly curated audio-driven facial animation subset. For each of these tasks, we determine the data distribution within the corresponding benchmark based on style diversity and task-specific requirements, and manually select the most suitable test cases. A detailed account of our data cleaning and dataset construction process can be found in Section 3. With hierarchical, thorough annotations, MagicAnime is tested to support multiple multimodal tasks, and can be tailored to downstream research and applications in cartoon animation video generation.

In summary, we make the following contributions:
\begin{itemize}
    \item We proposed MagicAnime, a multimodal \textbf{dataset and benchmarks} tailored for cartoon animation generation, incorporating video, audio, pose, and textual information, and supporting various video generation tasks.    
    \item We designed a hierarchical \textbf{annotation method} for data selection, multi-step processing, and high-quality multimodal labeling.  
    \item We validated the limitations of existing video generation methods on cartoon animation generation and improvements after finetuning with our MagicAnime dataset, showcasing its effectiveness and providing insights for relevant domains.
\end{itemize}

%% file: sec/2_related_work.tex
\begin{table*}[ht!]
    \centering
    \caption{MagicAnime dataset-subsets and benchmark size overview. 'All' represents the total amount of data in the subset and includes data from the respective benchmark. We detail the distribution of data for each of the three animation styles, American0, China, Japan, in each subset and its corresponding benchmark set. Keypoints, Audio, Text, represents the annotations for the three modalities contained in our dataset, and how many such annotations are contained in each subset.}
    \resizebox{\textwidth}{!}{
        \begin{tabular}{cc|cccc|ccc}
            \toprule
            \textbf{Supported Task} & \textbf{Subset} & \textbf{Total (clips)} & \textbf{American} & \textbf{China} & \textbf{Japan} & \textbf{Keypoints} & \textbf{Audio} & \textbf{Text} \\
            \midrule
            \multirow{2}{*}{\makecell{Audio-driven\\facial animation}} & All & 3k & 364 & 587 & 2.05k & 3k & 3k & None \\
            & Benchmark & 100 & 28 & 13 & 59 & 100 & 100 & None \\
            \midrule
            \multirow{2}{*}{\makecell{Face\\reenactment}} & All & 12.08k & 599 & 1.35k & 10.13k & 12.08k & None & None \\
            & Benchmark & 100 & 28 & 13 & 59 & 100 & None & None \\
            \midrule
            \multirow{2}{*}{\makecell{Image-to-video \\ generation}} & All & 400k & 90k & 60k & 250k & None & None & None \\
            & Benchmark & 300 & 100 & 50 & 150 & None & None & None \\
            \midrule
            \multirow{2}{*}{\makecell{Pose-driven \\ animation}} & All & 50k & 15k & 10k & 25k & 50k & None & 50k \\
            & Benchmark & 300 & 100 & 50 & 150 & 300 & None & 300\\
            \midrule
            \multirow{2}{*}{\makecell{Text-to-video \\ generation}} & All & 50k & 15k & 10k & 25k & 50k & None & 50k \\
            & Benchmark & 300 & 100 & 50 & 150 & 300 & None & 300\\
            \bottomrule
        \end{tabular}
    }
    \label{tab:dataset}
\end{table*}

\section{Related Work}
\label{gen_inst}
The field of video generation has garnered significant attention due to its wide range of applications~\cite{ma2025controllable}. Cartoon animation generation, as an important sub-task of video generation and visual art creation, has played an important role in entertainment, virtual avatars, and digital storytelling~\cite{wang2025diffusion}. In this section, we review recent work in video generation models, cartoon animation generation methods, and cartoon animation datasets.

\subsection{Video Generation Models}
With the advancement of generative modeling, large-scale video models have undergone significant transformation, particularly in diffusion-based frameworks. These video generation models often derive from the Stable Diffusion architecture~\cite{rombach2022high}. The UNet~\cite{ronneberger2015u}, originally employed for image generation tasks~\cite{song2020denoising}, has been modified to accommodate video generation by incorporating temporal aspects. VDM~\cite{ho2022video} adapts the 2D U-Net into a 3D framework, while another approach~\cite{guo2023animatediff} integrates 1D temporal attention with 2D spatial attention blocks to enhance computational efficiency. In particular, diffusion transformers (DiT~\cite{peebles2023scalable}), which are constructed solely using transformer blocks, outperform UNet in the generation of visual content~\cite{chen2024pixartalpha}. This architecture has been adapted for video models~\cite{hong2022cogvideo}, resulting in two prevalent variants: the original DiT employs cross-attention for text embeddings~\cite{feng2025dit4edit}, whereas MM-DiT~\cite{sd3} concatenates text embeddings with visual embeddings to enable comprehensive attention processing. In the realm of autoencoders, earlier methods~\cite{rombach2022high} utilized the standard VAE~\cite{kingma2013auto}, but recent iterations such as VQ-VAE~\cite{van2017neural} and VQGAN~\cite{esser2021taming} have refined model architecture to enhance reconstruction and compression capabilities. LTX-Video~\cite{hacohen2024ltx} adapts the VAE decoder for the final denoising phase, transforming latents into pixel data and generating missing high-frequency details during decoding. The text encoder is pivotal for text-based video creation. Leading video generation models predominantly use the T5 series~\cite{raffel2020exploring} as the primary text encoder, often paired with CLIP~\cite{radford2021learning}. In HunyuanVideo~\cite{kong2024hunyuanvideo}, T5 is replaced by a multimodal large language model to achieve a more robust alignment between textual and visual features. Concurrently, various studies, including~\cite{blattmann2023stable, Sora, hong2022cogvideo}, have shown promising outcomes in general video generation.

\subsection{Cartoon Animation Generation}
The generation of cartoon content concerns colorization~\cite{zhang2025magiccolor}, storytelling~\cite{wang2025magicscroll}, film production~\cite{song2023expanded} and many other subtasks. For example, AniDoc utilizes video diffusion models to automate the processes of colorization in 2D animations~\cite{meng2024anidoc}, whereas ToonCrafter~\cite{xing2024tooncrafter} effectively manages exaggerated non-linear motions and occlusions characteristic of cartoons. Numerous studies concentrate on the animation of cartoon characters~\cite{hu2023animate, zhu2024champ, wang2024humanvid, chen2024echomimic}. These studies introduce ReferenceNet~\cite{hu2023animate} to integrate spatial body information, facilitating the creation of target poses and bridging the gap between source and target poses. While these methods primarily focus on regulating body movements, several techniques emphasize facial movement control, collectively forming a comprehensive framework for character animation. Recent research~\cite{song2021talking, song2024adaptive, song2020denoising, liu2024emo} involves animating faces in a photo-realistic or toonified style using 3D Gaussian splatting or neural representations. For instance, TextToon~\cite{song2024texttoon} and Emo-Avatar~\cite{liu2024emo} leverage 3D Gaussian representations combined with large language models (LLMs) to adapt facial cartoon styles through text-based modifications. Ada-TalkingHead~\cite{song2024adaptive} uses neural keypoints to drive head animations, while Editable-Head~\cite{song2021talking} generates head animations with explicit landmark-based representations. Compared to the broader approach of body movement control, these techniques offer more refined solutions, delivering greater sophistication and intricacy in facial motion animation. Other tasks, such as inbetweening or interpolation, have also been extensively explored~\cite{xing2024tooncrafter, shen2022enhanced}.

\subsection{Datasets for Cartoon Animation Generation}
It has been noted that the quality of generative model outputs tends to improve with access to extensive datasets for training. Video data, especially domain-specific content, plays a crucial role in the development of generation models. However, acquiring high-quality animation video data poses more challenges than obtaining natural video datasets. To address this, several datasets have been introduced to bolster the effectiveness of video generation models. Earlier studies have released animation-related datasets such as ATD-12K~\cite{siyao2021deep} and AVC~\cite{wu2022animesr}. While these datasets, derived from various animation films, are beneficial for tasks like video interpolation and super-resolution, their small size is a notable limitation. Recently, Sakuga-42M~\cite{pan2024sakuga} has been introduced, featuring 1.2 million clips, marking an improvement over earlier datasets with only a few hundred clips. Nonetheless, this remains inadequate for training advanced video generation models, especially when compared to more comprehensive datasets like Panda-70M~\cite{chen2024panda} and InternVid-200M~\cite{wang2023internvid}. Furthermore, most clips in Sakuga-42M are low-resolution and shorter than 2 seconds, which limits the generation of high-quality videos. AniSora~\cite{jiang2024exploring} presents a dataset focused on cartoon animation generation, but it only includes textual labels, restricting its utility for detailed tasks such as audio-guided video generation and facial animation.

%% file: sec/3_method.tex
\begin{figure*}[t!]
    \centering
    \includegraphics[width=1\textwidth]{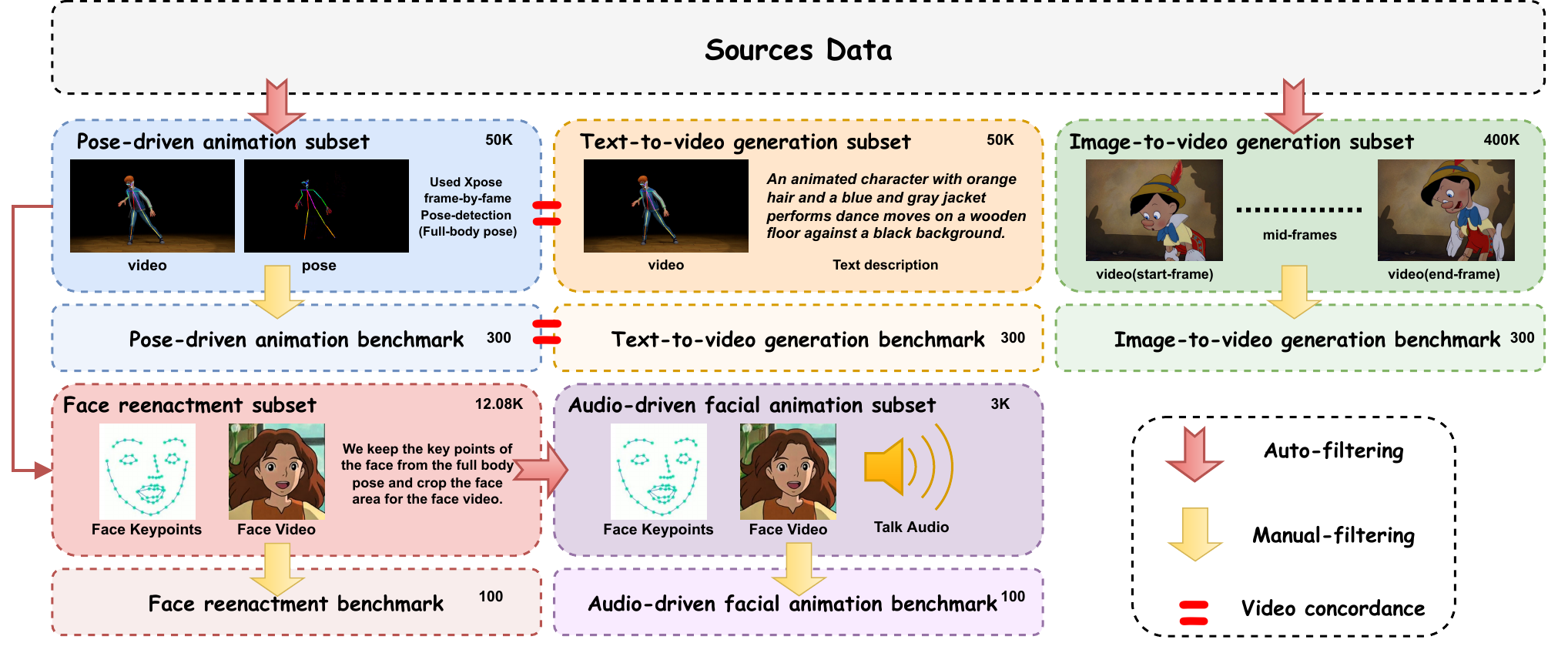}
    \caption{Dataset construction pipeline and overviews. The overall framework and screening process of our dataset is hierarchical The key innovation of the framework is the use of keypoint data to extract and filter video clips at different scales, thus generating full-body and facial video clips while completing the annotation. The red arrows shown in the figure represent automated screening and the yellow arrows represent manual screening. The equal sign indicates that the two subtasks have the same video data content in the subset. In part\ref{The MagicAnime Dataset}{} we will detail how to do automatic and manual filtering, and the characteristics of each subset.}
    \label{fig:pipeline}
    \vspace{-0.35cm}
\end{figure*}

\section{The MagicAnime Dataset}
\label{The MagicAnime Dataset}


\subsection{Data Sources}


\begin{figure*}[ht!]
    \centering
    \begin{minipage}{0.48\textwidth}
        \centering
        \includegraphics[width=\textwidth]{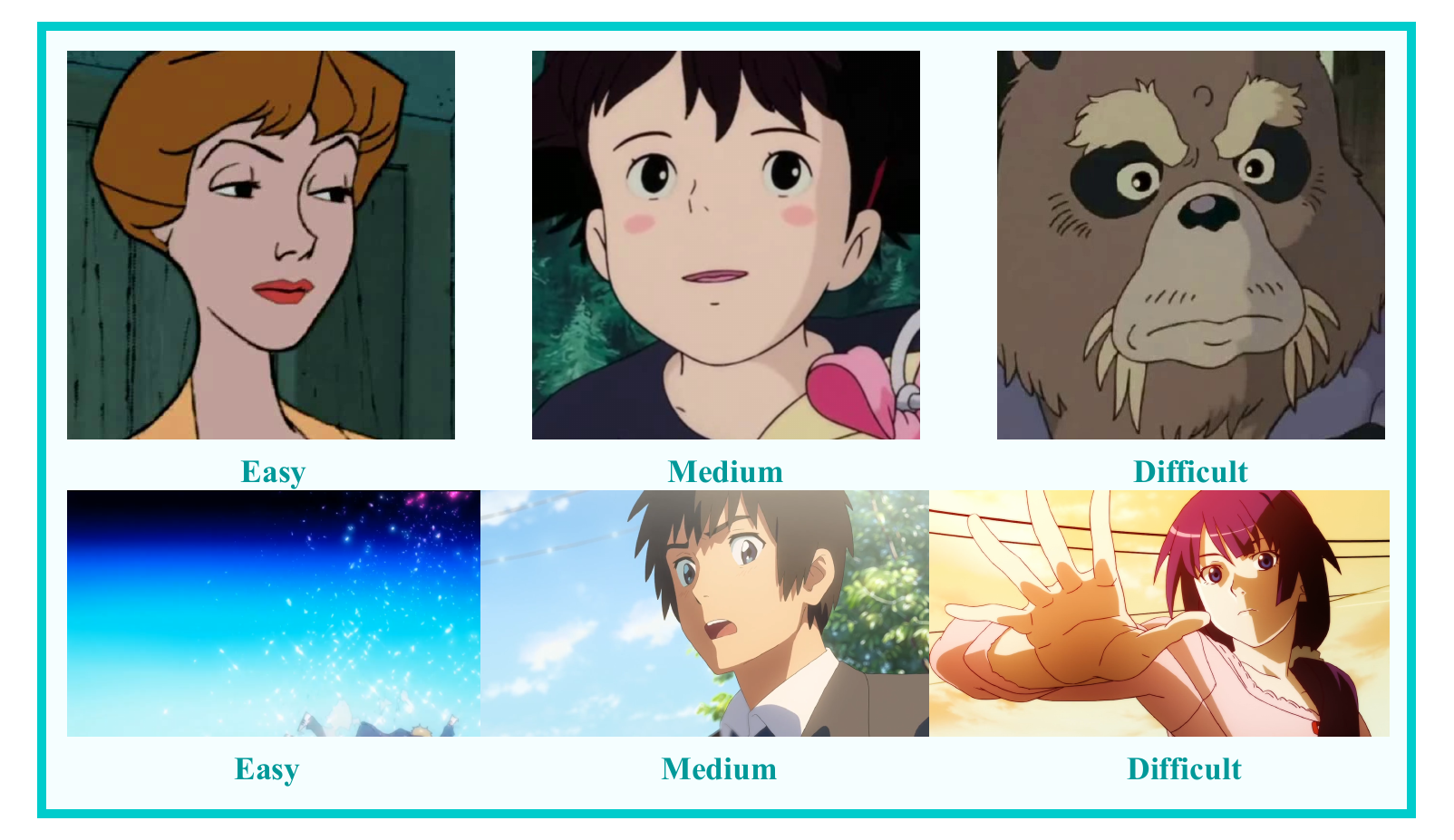}
        \caption{Examples of different levels of difficulty in our cartoon animation dataset. In the Portrait data category, there are simple data types that feature realistic drawings, cartoon characters of medium difficulty, and anthropomorphic animal characters of the highest difficulty. The data supported by Image2Video tasks are differentiated according to the range of motion of the screen and the complexity of the screen.}
        \label{fig:sub0}
    \end{minipage}
    \hfill
    \begin{minipage}{0.48\textwidth}
        \centering
        \vspace{-0.8cm}
        \includegraphics[width=\textwidth]{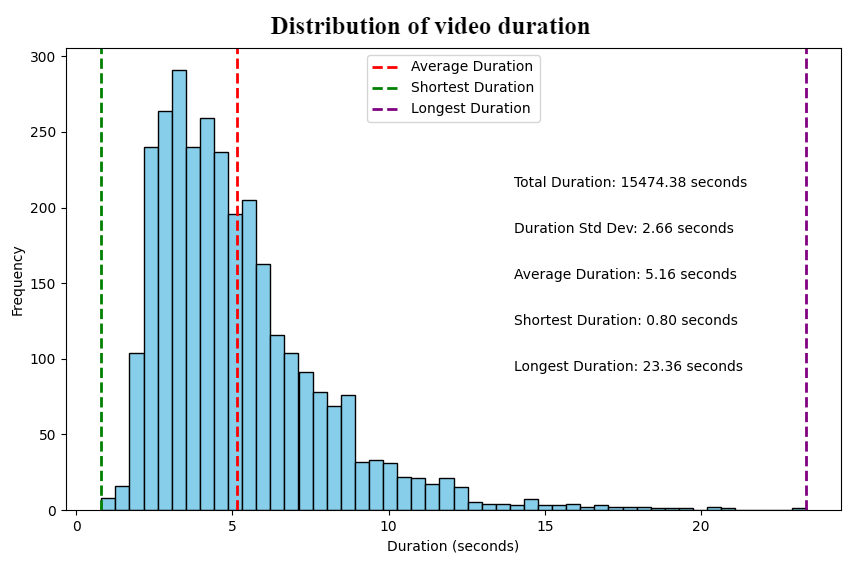}
        \caption{Distribution of video duration. We have kept the average length of video clips to around five seconds.}
        \label{fig:sub1}
    \end{minipage}
    \vspace{-0.35cm}
\end{figure*}

As shown in Figure 2, our dataset primarily consists of animated films from three major regional styles: 51 American animated films, 66 Japanese animated films (including 33 animated series), and 20 Chinese animated films. Based on our industry research in the animation domain, these three categories dominate the global animation market. Specifically, Japanese animation accounts for approximately 60\% of global production, while American animation constitutes around 30\%, and Chinese animation comprises less than 10\%, with the remaining share distributed among other regions. To ensure diversity in artistic styles and to reflect the real-world distribution of the animation market, we adopted a similar proportional distribution when constructing our dataset. This approach also facilitates further exploration of the performance differences and stylistic characteristics of various animation styles in AI-driven video generation.

We strived to ensure that at least 60\% of the videos in the dataset have a resolution of 1080P or higher. As shown in Figure 5, the majority of video clips in our dataset have a duration of approximately 5 seconds, while the number of frames follows the source videos. Since animated videos generally have lower frame rates, altering the frame rate may compromise the original video quality by distorting character motion speeds and deviating from the original design. Additionally, it should be noted that approximately 40\% of the source videos in our data set originate from the open source dataset Sakuga-42M~\cite{pan2024sakuga}.

\subsection{Dataset Statistics}
As shown in Table 1 and Figure 2(a), the MagicAnime dataset is composed of five subsets, each characterized by distinct modalities and annotations, designed to support various types of video generation tasks. Moreover, each subset includes a benchmark specifically designed for its corresponding modality and supported task, with carefully curated samples of varying scales.\\

\textbf{Audio-driven facial animation subset.} The first subset is specifically designed for facial animation generation tasks where audio serves as the control condition. This subset contains a total of 3K video clips, distributed as follows: 364 clips in the American animation style, 587 in the Chinese animation style, and 2.05K in the Japanese animation style. This distribution aligns with the overall proportion of animation styles in our dataset. The subset includes annotations for two modalities: (I) 68 facial keypoints for precise facial movement tracking and (II) speech audio that is consistent with the facial identity. All video clips are in a 512×512 rectangular format, ensuring that the character’s face is centered and occupies the majority of the frame. Additionally, all audio files are in MP3 format and have undergone speech enhancement and background noise reduction processing.\\

\textbf{Audio-driven facial animation benchmark.} The benchmark set consists of 100 carefully curated cases, all manually selected. The distribution remains consistent with the overall dataset proportions, with 28 cases in the American animation style, 13 in the Chinese style, and 59 in the Japanese style. Additionally, we ensured a reasonable allocation of difficulty levels among the cases to effectively evaluate the generation capabilities of different models, as illustrated in Figure 4. For the Audio-Driven Facial Animation task, we observed that the difficulty of a case is correlated with the \textit{degree of anthropomorphism} of the character's facial features. Here, "degree of anthropomorphism" refers to how closely an animated character's facial structure resembles human-like features. As shown in the last column of the first row in Figure 4, one of the most challenging cases is a \textit{fully bear-like anthropomorphic character}. Current models struggle to generate high-quality results for such cases, with some even failing completely. The primary reason is that the face perception models integrated into current methods fail to process such anthropomorphic characters, as they deviate significantly from human facial structures. Moreover, these cases fall outside the distribution of existing training data used in the current models.\\

\textbf{Face reenactment subset.} The Face Reenactment task involves generating facial animations driven by a sequence of face keypoints or face landmarks, similar to the Audio-Driven Facial Animation task but without the audio control condition. Due to the absence of the strict video-audio alignment requirement, this subset allows for a significant increase in data volume. The Face Reenactment Subset consists of 12.08K animated clips, distributed as follows: 599 clips in the American animation style, 1.35K in the Chinese style, and 10.13K in the Japanese style. In terms of annotation modalities, this subset retains facial keypoints annotations. The video clip dimensions remain consistent with those in the Audio-Driven Facial Animation Subset.\\

\textbf{Face reenactment benchmark.} In this task subset, the Face Reenactment benchmark shares the same benchmark set as the Audio-Driven Facial Animation task, as the only difference is the absence of the audio modality. Since both tasks evaluate the quality of generated videos using the same criteria, the benchmark remains consistent across them.\\

\textbf{Image-to-video generation subset.} The Image-to-Video Generation (I2V) task is a fundamental problem in video generation. Our I2V data set also supports a related task, video frame interpolation. Following the data processing paradigm of SVD{}, we have curated and structured this subset accordingly.  This subset contains a total of 400K video clips, distributed as follows: 90K in the American animation style, 60K in the Chinese style, and 250K in the Japanese style. It is important to note that due to computational resource limitations, we have not yet completed the text annotations for this subset. However, users can refer to the data processing methods of EasyAnimate and CogVideo to independently annotate text for training purposes.\\

\textbf{Image-to-video generation benchmark.} The benchmark for the Image-to-Video Generation subset follows the same proportional distribution strategy. This benchmark consists of a total of 300 video clips, distributed as follows: 100 in the American animation style, 50 in the Chinese style, and 150 in the Japanese style. The difficulty levels of the benchmark cases are illustrated in the second row of Figure 4. The selection of cases was performed manually, primarily considering two factors: the complexity of the visual content and the speed and complexityy of the motion within the scene.\\

\textbf{Pose-driven animation subset.} The Pose-Driven Animation task refers to character animation generation driven by pose sequences, with a primary focus on motion. This subset contains a total of 50K video clips, distributed as follows: 15K in the American animation style, 10K in the Chinese style, and 25K in the Japanese style. The dataset includes annotations in two modalities: frame-by-frame annotation of 133 full-body keypoints and textual descriptions that emphasize the character's movements within the video.\\

\textbf{Pose-driven animation benchmark.} The benchmark subset for this task consists of a total of 300 video clips, distributed as follows: 100 in the American animation style, 50 in the Chinese style, and 150 in the Japanese style. To ensure the accuracy and smooth alignment of pose annotations with the characters, all benchmark cases were manually selected. The subset includes video clips featuring both upper-body and full-body scenes, providing a diverse range of pose-driven animation scenarios.\\

\textbf{Text-to-video generation subset.} The Text-to-Video Generation (T2V) task is a fundamental problem in video generation. Given the existence of larger-scale T2V animation datasets such as Anisora and Sakuga, our T2V subset is designed with a different focus. Specifically, our data set places greater emphasis on character motion. During the refinement of annotation prompts, we prioritized enhancing motion descriptions to better capture dynamic aspects of animation. This subset aligns with the Pose-Driven Animation task and serves both as its textual annotation and as an independent Text-to-Video Generation dataset. Consequently, the benchmark for this subset is identical to that of the Pose-Driven Animation task and will not be further elaborated here.
\subsection{Data Processing}
Data processing is the fundamental and most important part of this work. Because our dataset is multimodal and contains multiple subsets, the whole data processing process has many tedious steps and is similar to that of other video datasets. Therefore, we will focus on the data processing steps related to the cartoon animation characteristics of our dataset and the multimodality in this section, and we will focus on the two aspects of data cleaning, hierarchical multimodal data filtering and annotation. Expand on our data processing steps.
\subsubsection{Data cleaning}
To efficiently process the raw video data, we utilized the PySceneDetect library~\cite{castellano2022pyscenedetect}, applying its ContentDetector to perform scene boundary detection. This enabled us to segment long animation movies and videos into shorter clips while preserving scene consistency. Most existing video data set construction pipelines typically employ methods such as CLIP~\cite{radford2021learning} or DINOv2~\cite{oquab2023dinov2} to compute the framewise similarity between keyframes or the starting and end frames. These approaches help filter out clips with abrupt scene transitions caused by over- or under-segmentation from PySceneDetect, and they are generally effective for natural video data. However, cartoon animation videos present unique challenges due to their vibrant color dynamics and stylized scene transitions, which can significantly reduce the accuracy of automated segmentation. To address this, we implemented a secondary refinement step based on pose annotations obtained in the downstream pipeline. Specifically, we retained only those video segments that contain a single human character, as required by the pose-driven animation subset. Furthermore, to ensure scene consistency and a stable presence of the target character, we filtered out temporal intervals where no valid pose sequence was detected over a sufficient number of consecutive frames. Although this process inevitably led to the exclusion of a large number of clips, it effectively aligned with the objectives of the pose-driven animation task and ensured a high level of data usability in the filtered subset.

\subsubsection{Hierarchical multimodal data filtering and annotation}
Our goal is to provide a large-scale, multi-modal dataset capable of supporting various cartoon animation generation tasks. To this end, I have designed a hierarchical data processing framework for filtering and annotation, tailored to different modalities and data types. This framework enables complementary selection across modalities and supports simultaneous annotation and filtering, as show in Figure \ref{fig:pipeline}

First, as previously mentioned, we performed large-scale human pose keypoint detection. During the annotation of human keypoints, we simultaneously filtered out video clips containing characters, thereby constructing our first pose-guided subset. Next, we annotated the video clips in the pose-guided subset with textual descriptions focusing on character actions, in order to build the corresponding text-to-video subset. In parallel, more video clips were annotated with general textual descriptions, and their first frames were extracted to construct the image-to-video subset.

For the face reenactment and audio-driven facial animation subsets, we first used the 68 facial landmarks from Xpose to detect face bounding boxes and cropped single-character face video clips with consistent identity, resulting in square videos of size 512×512. Side-profile videos were filtered out based on the bounding box calculated from facial landmarks. This allowed us to construct a high-quality face reenactment subset.

Finally, for the most fine-grained audio-driven facial animation subset, we further filtered the face reenactment data to meet stricter criteria: (1) valid corresponding audio must be available; (2) the audio must be temporally aligned with the character’s lip movements in the video; (3) the audio must not contain background noise or speech from other characters.
Additionally, we applied denoising and background separation to the audio files, retaining only the character’s clean speech. Due to the stringent requirements, the size of the final audio-driven facial animation subset is relatively small. However, it maintains high quality and is well-suited for training or finetuning existing baseline methods.

%% file: sec/4_evaluation.tex
\section{Experiments}

\begin{table*}[ht!]
    \centering
    \caption{Performance Comparison of Different Models. VSR: Valid Sample Ratio. Please refer to Sec.~\ref{sec:procedure} for the detailed descriptions of annotations in the parentheses. \textit{Before:} before finetuning. \textit{After:} after finetuning. \textit{Kpt\_1:} the original unipose keypoints as input. \textit{Kpt\_2:} our human-curated keypoints as input. \textit{Audio\_1:} audio with modified tone as input. \textit{Audio\_2:} audio with random voice as input.}
    \resizebox{\textwidth}{!}{
        \begin{tabular}{ccccccc}
            \toprule
            \textbf{Task} & \textbf{Method} & \textbf{VSR} & \textbf{PSNR↑} & \textbf{SSIM↑} & \textbf{LPIPS↓} & \textbf{L1-loss↓} \\
            \midrule
            \multirow{5}{*}{\makecell{Audio-driven \\ facial animation}} 
            & Echomimic~\cite{chen2024echomimic} & 20\% & 14.787 & 0.460 & 0.403 & 0.119 \\
            & AniPortrait~\cite{wei2024aniportrait} (\textit{before}) & 40\% & 15.780 & 0.500 & 0.349 & 0.107 \\
            & AniPortrait (\textit{after}) & 40\% & 16.251 & 0.525 & 0.323 & 0.099 \\
            & AniPortrait (\textit{audio\_1}) & 40\% & 16.229 & 0.529 & 0.328 & 0.099 \\
            & AniPortrait (\textit{audio\_2}) &  40\% & 16.354 & 0.535 & 0.321 & 0.097 \\
            & Hallo~\cite{xu2024hallo} & 55\% & 18.005 & 0.586 & 0.245 & 0.081 \\
            \midrule
            \multirow{7}{*}{\makecell{Face reenactment}} 
            & Moore-AnimateAnyone~\cite{MooreThreads_Moore_AnimateAnyone} & 63\% & 13.800 & 0.417 & 0.442 & 0.140 \\
            & FADM~\cite{zengbohan0217_FADM} & \textbf{100}\% & 16.620 & 0.556 & 0.359 & 0.116 \\
            & LivePortrait~\cite{guo2024liveportrait} (\textit{kpt\_1}) & \textbf{100}\% & 17.250 & 0.573 & 0.301 & 0.094 \\
            & LivePortrait (\textit{kpt\_2}) & \textbf{100}\% & 17.269 & 0.574 & 0.300 & 0.094 \\
            & Follow-Your-Emoji~\cite{ma2023follow} & 51\% & 19.074 & 0.632 & 0.224 & 0.091 \\
            & AniPortrait (\textit{before}) & 54\% & \textbf{24.648} & \textbf{0.920} & \textbf{0.067} & \textbf{0.023} \\
            & AniPortrait (\textit{after}) & 54\% & \textbf{24.648} & \textbf{0.920} & \textbf{0.067} & \textbf{0.023} \\
            \midrule
            \multirow{2}{*}{\makecell{Image-to-video \\ generation}} 
            & DynamiCrafter~\cite{xing2023dynamicrafter} & 97\% & 30.678 & \textbf{0.738} & \textbf{0.189} & —\\
            & SVD~\cite{blattmann2023stable} & \textbf{99.5}\% & \textbf{31.162} & 0.735 & 0.191 & —\\ 
            \midrule
            \makecell{Frame interpolation}
            & ToonCrafter~\cite{xing2024tooncrafter} & 98\% & 30.547 & 0.717 & 0.210 & —\\
            \bottomrule
        \end{tabular}
    }
    \label{tab:experiment}
\end{table*}

\begin{figure}[ht!]
    \centering
    \includegraphics[width=1\columnwidth]{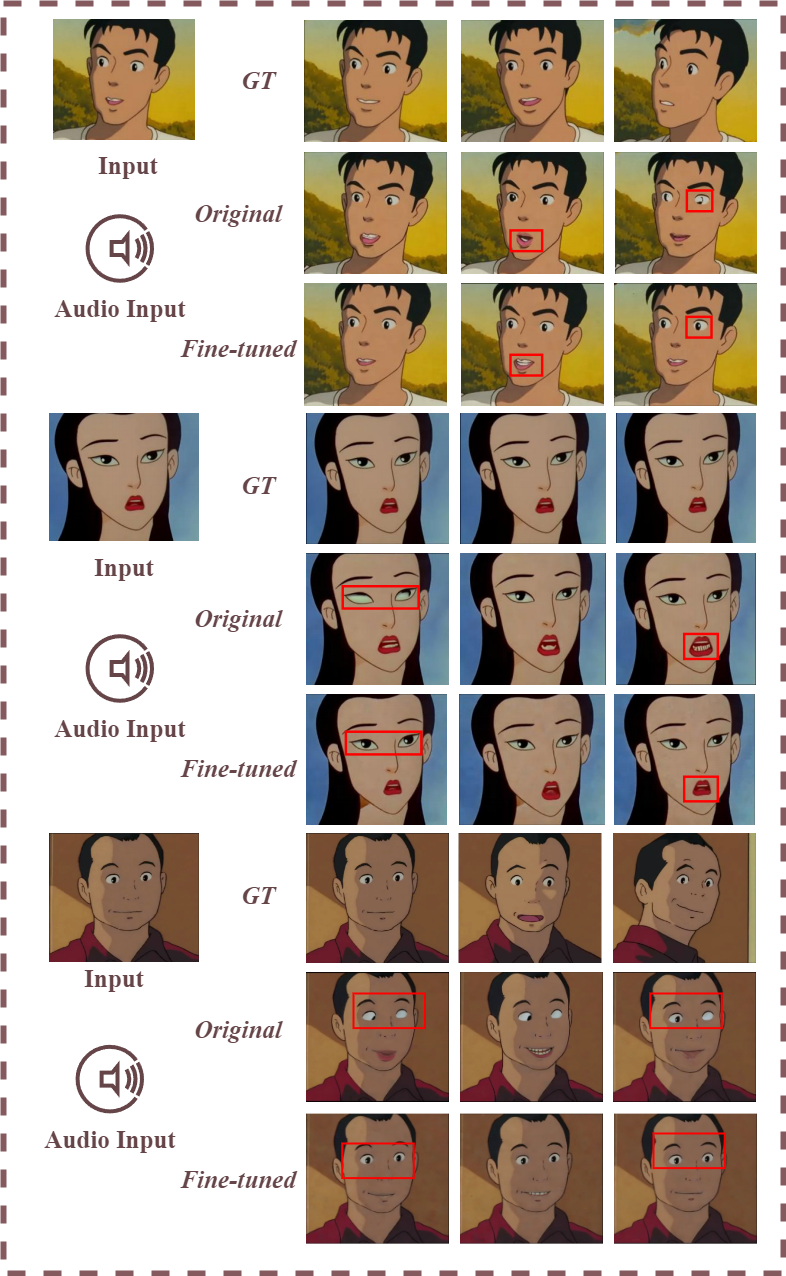}
    \caption{Comparison between ground truth (GT) videos, results generated by the original AniPortrait, and its finetuned model. Finetuning the approach on our dataset helps the model generate more detailed cartoon portrait features, including stylistic eyes, lips, teeth, nose, etc.}
    \label{fig:tuning}
    \vspace{-0.35cm}
\end{figure}

\subsection{Task and Benchmark}
To assess the effectiveness of our MagicAnime dataset on baseline methods, we conducted a series of qualitative evaluations across various generative tasks using our established benchmark. As shown in Table~\ref{tab:experiment}, we mainly included the following tasks: audio-driven facial animation, face reenactment, image-to-video
generation, and frame interpolation. According to Table~\ref{tab:dataset}, our benchmark consists of 100 video clips (with audio) for audio-driven facial animation and face reenactment generation and 300 video clips for image-to-video generation and frame interpolation. The clips used in the tasks cover a wide range of diverse countries, styles, and genres, with 12\% of them being non-humanoid samples to assess the generalizability of different methods.

\subsection{Baselines and Metrics}
According to Table~\ref{tab:experiment}, we selected a series of different models for each task and compared their performance on the corresponding benchmarks. Specifically, for audio-driven facial animation, we selected Echomimic~\cite{chen2024echomimic}, Aniportrait~\cite{wei2024aniportrait}, and Hallo~\cite{xu2024hallo}; for face reenactment, we selected Moore-AnimateAnyone~\cite{MooreThreads_Moore_AnimateAnyone}, FADM~\cite{zengbohan0217_FADM}, LivePortrait~\cite{guo2024liveportrait}, Follow-Your-Emoji~\cite{ma2023follow}, and AniPortrait~\cite{wei2024aniportrait}; for image-to-video generation and frame interpolation, we selected ToonCrafter~\cite{xing2024tooncrafter}, DynamiCrafter~\cite{xing2023dynamicrafter}, and SVD~\cite{blattmann2023stable}.

According to Table~\ref{tab:experiment}, we included Valid Sample Ratio (VSR), Peak Signal-to-Noise Ratio (PSNR)~\cite{horé2010psnr}, Structural Similarity Index (SSIM)~\cite{wang2004ssim}, Learned Perceptual Image Patch Similarity (LPIPS)~\cite{zhang2018lpips}, and Least Absolute Deviations (L1-loss)~\cite{tibshirani1996lasso} as evaluation metrics. Among them, the VSR refers to the portion of results that can be successfully generated, which represents the adaptability of our benchmark to different models. Other quantitative metrics including PSNR, L1-loss, SSIM, and LPIPS were used to assess pixel-level and perceptual-level performance.

\subsection{Experimental Procedure}
\label{sec:procedure}
In addition to evaluating the performance of existing methods, we also finetuned the AniPortrait framework as a representative method on both audio-driven facial animation and face reenactment to evaluate the effectiveness of our dataset. We abbreviate these experimental groups in parentheses after the corresponding methods in Table~\ref{tab:experiment} and aim to validate the following hypotheses:

First, does finetuning a model with our dataset improve task performance? For this test, we selected AniPortrait~\cite{wei2024aniportrait}, denoted as \textit{before} for the original model and \textit{after} for the finetuned model. In finetuning the AniPortrait model, the objective was to optimize this module while keeping other model components fixed. We utilized 2,900 samples from the Train\_set in Table~\ref{tab:dataset}. The model was trained for 36,432 steps with a learning rate of $1 \times 10^{-5}$, resulting in effective convergence of the loss function.

Second, does curating unipose keypoints improve task performance? We chose LivePortrait~\cite{guo2024liveportrait} for this test, with the original unipose keypoints labeled as \textit{kpt\_1} and our human-curated keypoints as \textit{kpt\_2}.

Third, how do the tone and rhythm of driving audio affect the generated video? We also selected AniPortrait for this test. We modified the tone of the driving audio (labeled as \textit{audio\_1}) and generated random, meaningless human voice as driving audio (labeled as \textit{audio\_2}) for comparison.

\subsection{Quantitative Results}
From the results displayed in Table~\ref{tab:experiment}, we derived several observations and potential explanations: 

First, different models showed varied performance across the three tasks, with Hallo and AniPortrait excelling in audio-driven facial animation and face reenactment. The results also demonstrate that there are no apparent conflicts among the five metrics we selected. In addition, methods that demonstrate strong performance on one task may also exhibit favorable results on the other task.

Second, curating unipose keypoints can improve the performance of face reenactment compared to the original keypoints. This confirms that face reenactment performance is affected by the accuracy of keypoint annotations. However, the performance gain from the human curation process is not significant.

Third, the tone, content, and rhythm of the driving audio have minimal impact on audio-driven facial animation using the AniPortrait framework. Based on the final conclusion, we hypothesize that this may result from the typical animated film production process, where a large portion of anime characters' mouth shapes only manifest as mechanically regular opening and closing.

\subsection{Qualitative Results}
Figure~\ref{fig:tuning} showcases the qualitative comparison between ground truth video samples and the results generated by the original AniPortrait model and our finetuned model. We observed the following findings and insights: 

First, finetuning our dataset notably enhances the model's ability to preserve facial details such as eye features, lips, teeth, and nose. However, the model still exhibits limitations in adapting to overall head motion and orientation changes compared to the ground truth. 

Second, the model struggles to accurately reflect high-level audio features, such as emotions, in the generated video. This is evident in the lack of exaggerated expressions or motions that would typically correspond to the audio content. 

Third, generating videos from different keyframes yields significantly varied results, indicating a lack of generalizability in the baseline model. In future endeavors, we aim to improve generation outcomes through innovative model architectures and refined training strategies.

%% file: sec/5_conclusion.tex
\section{Conclusion}

In this work, we have successfully addressed the significant challenges in generating high-quality cartoon animations by constructing the MagicAnime dataset, a large-scale, multimodal, and multitasking dataset specifically designed for cartoon animation videos. The total volume of the video clips exceeds 400k. Our approach bridges the existing gap between real-world motion data and the stylized, hand-crafted nuances of cartoon animation. By developing an effective data processing framework, we resolved key issues such as whole-body and facial motion region filtering, facial video cropping with audio alignment, and implemented rigorous manual filtering to ensure data quality.

Our work highlights the scarcity and challenges of generating high-quality cartoon animations with existing datasets and methods, which often fail to capture the unique motion dynamics and stylistic characteristics of cartoons. Through benchmarking on the latest video generation frameworks, we have demonstrated that fine-tuning on our MagicAnime dataset significant improves animation quality, particularly in the portrait audio-to-video task. 
Our dataset holds significant potential for applications in various fields such as animation character recognition, detection, segmentation, parsing, action recognition, and video generation. In future work, we intend to extend the modalities to include textual annotations for the description of two-dimensional motion.

\textbf{Disclaimer:} 
We acknowledge the importance of copyright protection. The use of this dataset is restricted to research purposes only, and access to the dataset will be controlled in strict accordance with the application review process and signed agreements to ensure its legitimate use. Access is only available to applicants affiliated with research institutions. We will prioritize releasing the benchmark test set with references to the original animation, similar to previous work on the resolution enhancement of animated content films~\cite{wu2022animesr,siyao2021deep}.